\documentclass[]{article}
\usepackage[numbers,sort&compress]{natbib}
\usepackage{graphicx}
\usepackage{authblk}
\usepackage[colorlinks = true,
            linkcolor = blue,
            urlcolor  = blue,
            citecolor = blue,
            anchorcolor = blue]{hyperref}
\begin{document}

\title{A System for Accessible Artificial Intelligence}

\author[1]{Randal S. Olson}
\author[1,2]{Moshe Sipper}
\author[1]{William La Cava}
\author[1]{Sharon Tartarone}
\author[1]{Steven Vitale}
\author[1]{Weixuan Fu}
\author[1,3]{Patryk Orzechowski}
\author[1]{Ryan J. Urbanowicz}
\author[1]{John H. Holmes}
\author[1]{Jason H.~Moore}

\affil[1]{Institute for Biomedical Informatics, University of Pennsylvania, Philadelphia, PA 19104}
\affil[2]{Department of Computer Science, Ben-Gurion University, Beer-Sheva 8410501, Israel}
\affil[3]{Department of Automatics and Biomedical Engineering, AGH University of Science and Technology, al. Mickiewicza 30, 30-059 Krakow, Poland}

\date{}

\maketitle

\begin{abstract}
While artificial intelligence (AI) has become widespread, many commercial AI systems are not yet accessible to individual researchers nor the general public due to the deep knowledge of the systems required to use them. We believe that AI has matured to the point where it should be an accessible technology for everyone. We present an ongoing project whose ultimate goal is to deliver an open source, user-friendly AI system that is specialized for machine learning analysis of complex data in the biomedical and health care domains. We discuss how genetic programming can aid in this endeavor, and highlight specific examples where genetic programming has automated machine learning analyses in previous projects.
\end{abstract}

\section{Introduction}

A central goal of artificial intelligence (AI) is to use computational hardware and software to solve complex problems in a human-competitive manner~\cite{kannappan2015analyzing}. The practicality of this goal is that AI can be tasked with solving problems or performing functions that humans cannot perform or simply don't have time for. Most AI methodologies can be grouped into top-down approaches, wherein cognition is viewed as a high-level phenomenon that is independent of the low-level details, or bottom-up approaches, which define basic computational building blocks such as artificial neurons that collectively give rise to ``emergent''~\cite{ronald1999design} intelligent behavior. The top-down approach has been difficult to realize given the inherent complexity of human cognition. However, the bottom-up has had some success owing to the availability of sophisticated algorithms such as genetic programming (GP)~\cite{koza1992genetic} and deep neural networks~\cite{goodfellow2016deep}. This is particularly true today with abundant and inexpensive high-performance computing, leading to many human-competitive success stories~\cite{kannappan2015analyzing}.

Medical applications of AI have had a long history with both successes and failures. One of the early successes was the Mycin system, which was designed to predict the antibiotic that a patient with an infection should receive in the intensive care unit~\cite{buchanan1984rule}. Mycin combined a knowledge base along with a set of rules implemented as part of an expert system. The system was demonstrated to be human-competitive, but was never put into clinical practice because of legal concerns and the time it took clinicians to enter the patient data required for Mycin to make the predictions. The field of AI has matured since Mycin was developed and, importantly, computing power has grown tremendously in parallel. Examples of modern AI successes include IBM's Watson, which beat the world champion of the game show Jeopardy~\cite{ferrucci2012introduction}. The Watson AI system that won Jeopardy combined knowledge representation, information retrieval, natural language processing, and machine learning along with high-performance computing to access and exploit a knowledge base that included the Wikipedia text corpus. This was a milestone in AI because it showed that a computational system could compete with humans on difficult language processing tasks. Watson is now being marketed in the health care domain although the jury is still out on its effectiveness.

Commercial AI systems such as Watson show potential but are not yet accessible to individual researchers nor the general public due the cost and the complexity of working with a team from IBM. It is our working hypothesis that,
\begin{quote}
 \textit{\textbf{AI has matured to the point where it should be an accessible technology for everyone}}.
\end{quote}
Democratization of AI will be important if we seek to integrate this exciting new technology into multiple different domains, as demonstrated by recent efforts such as Orange~\cite{Demsar2013Orange}. We describe here the early development stages of an open source and user-friendly AI system---PennAI (\url{http://pennai.org})---for machine learning analysis of complex data in the biomedical and health care domains. We focus our initial efforts on the classification of biomedical endpoints such as disease susceptibility. We describe in turn below each of the components of our AI system and then end with an example and a discussion of how we envision this system being used to solve complex biomedical problems. Further, we discuss how GP can aid in enhancing PennAI, and highlight specific examples where GP has automated machine learning analyses in previous work.

\begin{figure}
\begin{center}
\includegraphics[width=0.9\textwidth]{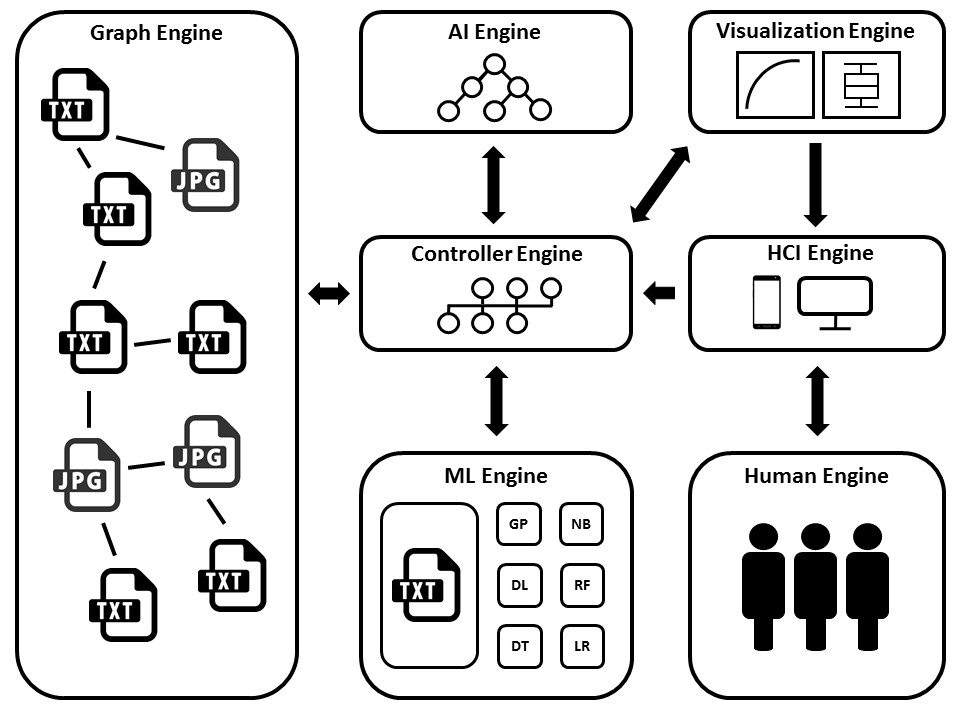}
\end{center}
\caption{The components of PennAI, a user-friendly AI system developed at the University of Pennsylvania.}
\label{fig-penn-ai}
\end{figure}

\begin{figure}
\begin{center}
\includegraphics[width=0.9\textwidth]{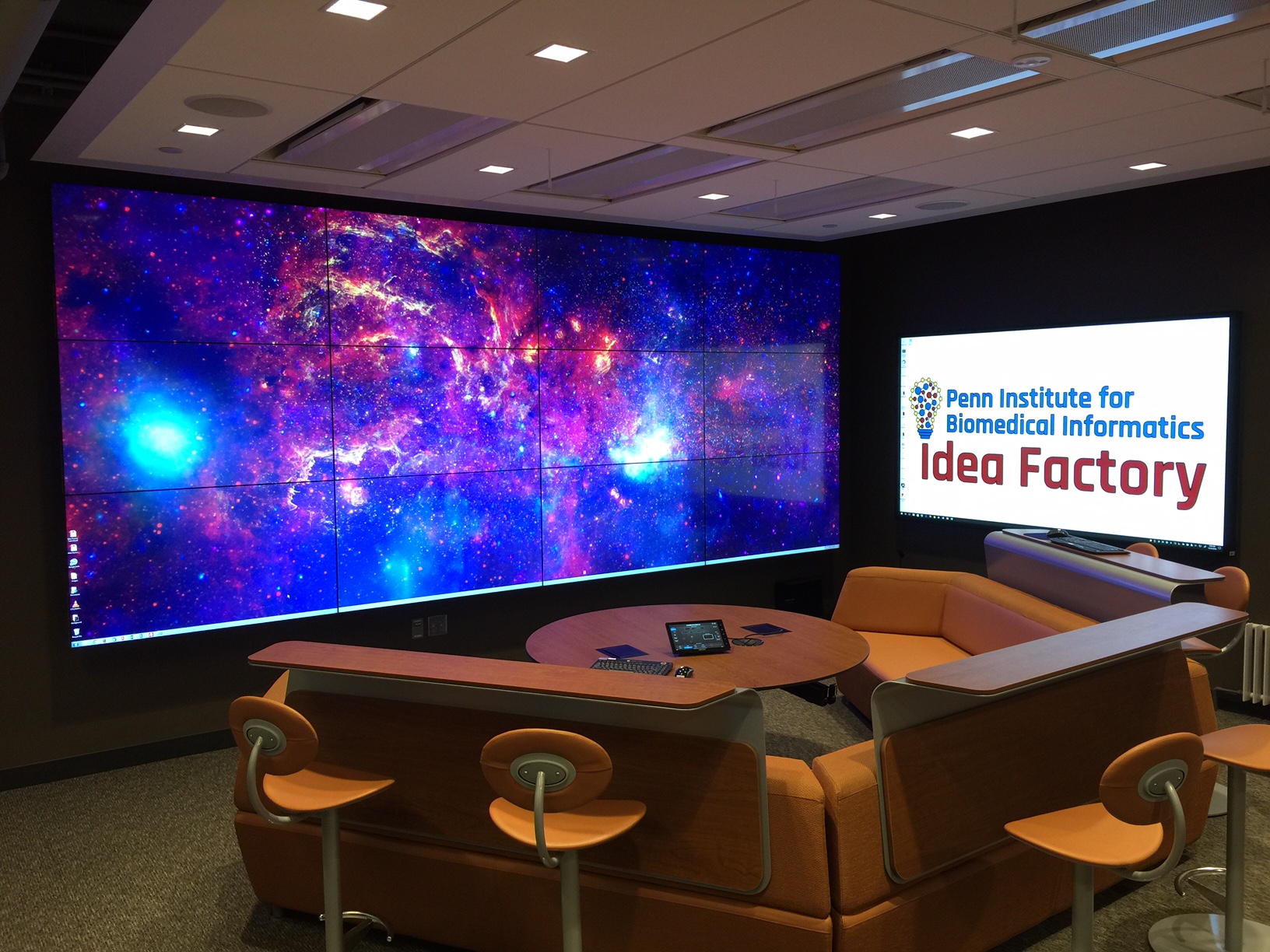}
\end{center}
\caption{The ``Idea Factory,'' home of PennAI.}
\label{fig-idea-factory}
\end{figure}

The components of PennAI include a human engine (i.e., the user); a user-friendly interface for interacting with the AI; a machine learning engine for data mining; a controller engine for launching jobs and keeping track of analytical results; a graph database for storing data and results (i.e., the memory); an AI engine for monitoring results and automatically launching or recommending new analyses; and a visualization engine to displaying results and analytical knowledge (Figure~\ref{fig-penn-ai}). This AI system provides a comprehensive set of integrated components for automated machine learning (AutoML), thus providing a data science assistant for generating useful results from large and complex data problems. 
PennAI is housed in the ``Idea Factory,'' a facility designed to facilitate collaboration and promote new methods of communicating and presenting scientific innovation. The Idea Factory makes sophisticated data visualization and artificial intelligence analytics easy for users across the entire Penn community (Figure~\ref{fig-idea-factory}).

\section{The Human Engine}

The most important component of the proposed AI system is the user. Contrary to some claims that AI will replace human users, we see the human as an integral part of the discovery process and a partner with the AI. One way to view this partnership is with the human as the driver of the discovery process and the AI as the data science assistant. Thus, the AI provides an additional set of hands in a modern data science discovery environment that might include human teammates with expertise in computer science, statistics, and applied mathematics. We have previously suggested this idea of human-computer interaction that places the human user at the epicenter~\cite{moore2007genome}. This idea has also previously been explored from the point of view of the user or domain expert~\cite{Langley02lessonsfor}.

Langley \cite{Langley02lessonsfor} provides five important tips that are relevant to thinking about the relationship humans have with AI for data mining using machine learning. First, traditional machine learning notations are not easily communicated to scientists. This consideration is important because a machine learning model may not be interpretable by a user. Second, scientists often have initial models that should influence the discovery process. Domain-specific knowledge can be critical to the discovery process. Third, scientific datasets are often rare and difficult to obtain. It often takes years to collect and process the data before it can be analyzed. As such, it is important that the analysis is carefully planned and executed, and that any general feedback about the performance the learning process is not lost between studies. Fourth, scientists want models that move beyond description and provide explanations of the data. Explanation and interpretation are paramount to the user. Finally, scientists want computational assistance rather than a complete replacement of themselves. Langley~\cite{Langley02lessonsfor} further suggests that users want interactive discovery environments that help them understand their data while at the same time giving them control over the modeling process. Collectively, these five lessons suggest that synergy between the user and the AI is critical. With this in mind, our proposed AI system includes a graphical user interface (GUI) that allows the user to easily launch analyses, view the results, and give the AI feedback about what results are useful or interesting.

\section{The Human-Computer Interaction Engine}

\begin{figure}
\begin{center}
\includegraphics[width=\textwidth]{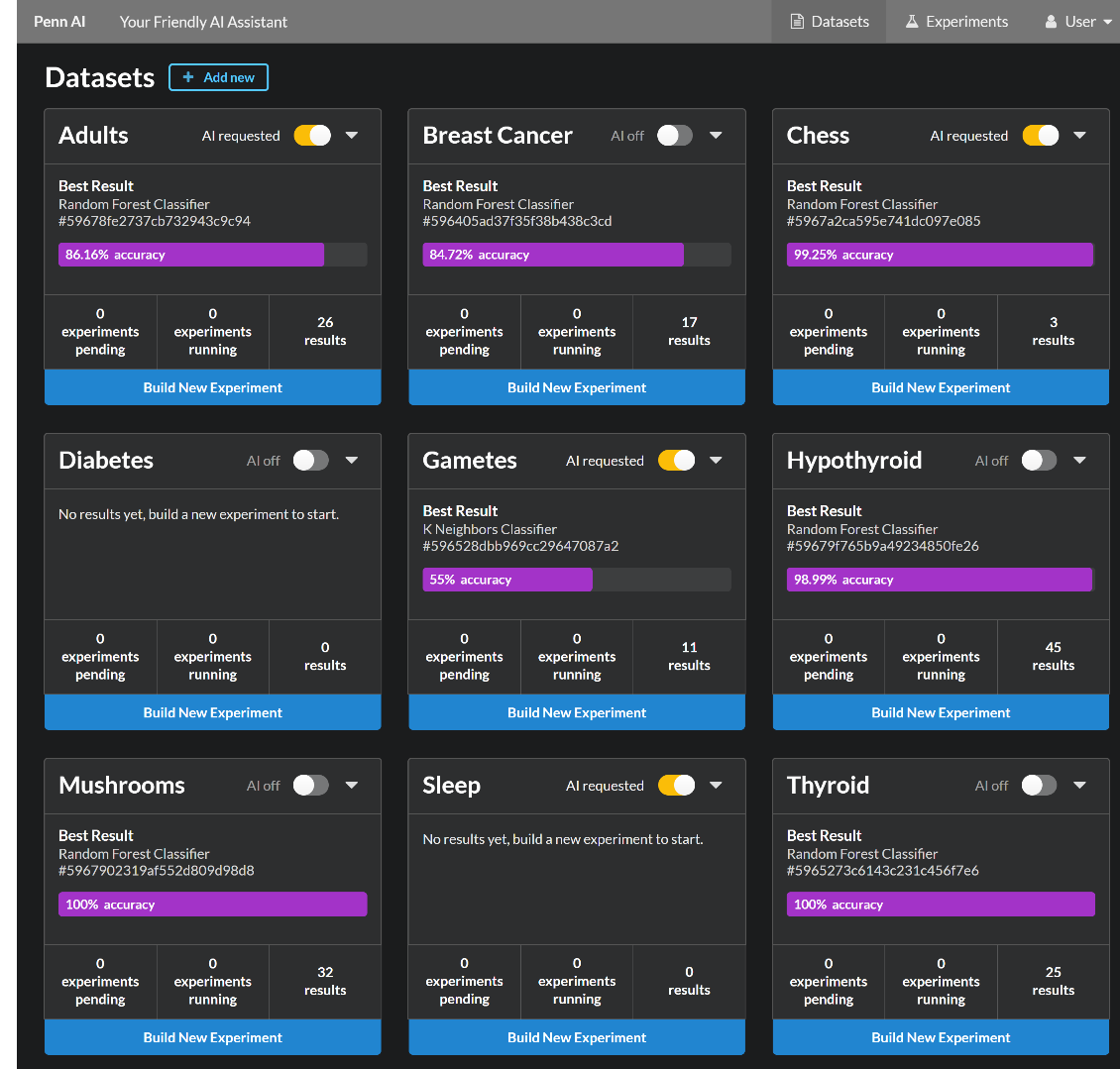}
\end{center}
\caption{Prototype of the graphical user interface for managing and viewing datasets.}
\label{fig:prototype-dataset-management}
\end{figure}

As described above, a key component of PennAI is human-computer interaction. The first important feature is to make it easy for the user to directly launch machine learning analyses by choosing a method and its parameter settings from an intuitive push-button menu implemented through the web using JavaScript. The user can launch single analyses or, in an advanced mode, launch a grid search across multiple methods and parameter settings. The methods and the controller that keep track of these analyses is described below. Figures~\ref{fig:prototype-dataset-management} and~\ref{fig:prototype-ml-gui} show prototypes of our GUI for uploading and viewing datasets for analysis and launching machine learning analyses on those datasets, respectively. Our JavaScript implementation is compatible with mobile devices, which allows the user to interact with the AI system from any Internet-connected device.

The second key feature of PennAI is the ability to toggle the AI on and off for automated analysis, shown in Figure~\ref{fig:prototype-dataset-management}. An AI toggle allows the user to turn the AI on and set parameters controlling the maximum number of runs the AI can launch, as well as the frequency of updates the user would like to receive by email or text message. The GUI also provides a simple thumbs up/down selection for each result received by PennAI, which provides feedback to PennAI that is incorporated into its expert knowledge system.

\section{The Machine Learning Engine}

Our first application of PennAI is for data mining using machine learning in the biomedical domain. Here, we make use of an extensive open source machine learning library in Python called scikit-learn~\cite{pedregosa2011scikit}. Scikit-learn provides peer-reviewed implementations of several common supervised and unsupervised machine learning algorithms, data preprocessing methods, feature engineering and selection methods, hyperparameter optimization procedures, and more. To most users, scikit-learn is considered to be the standard machine learning library in Python.

\begin{figure}
\begin{center}
\includegraphics[width=\textwidth]{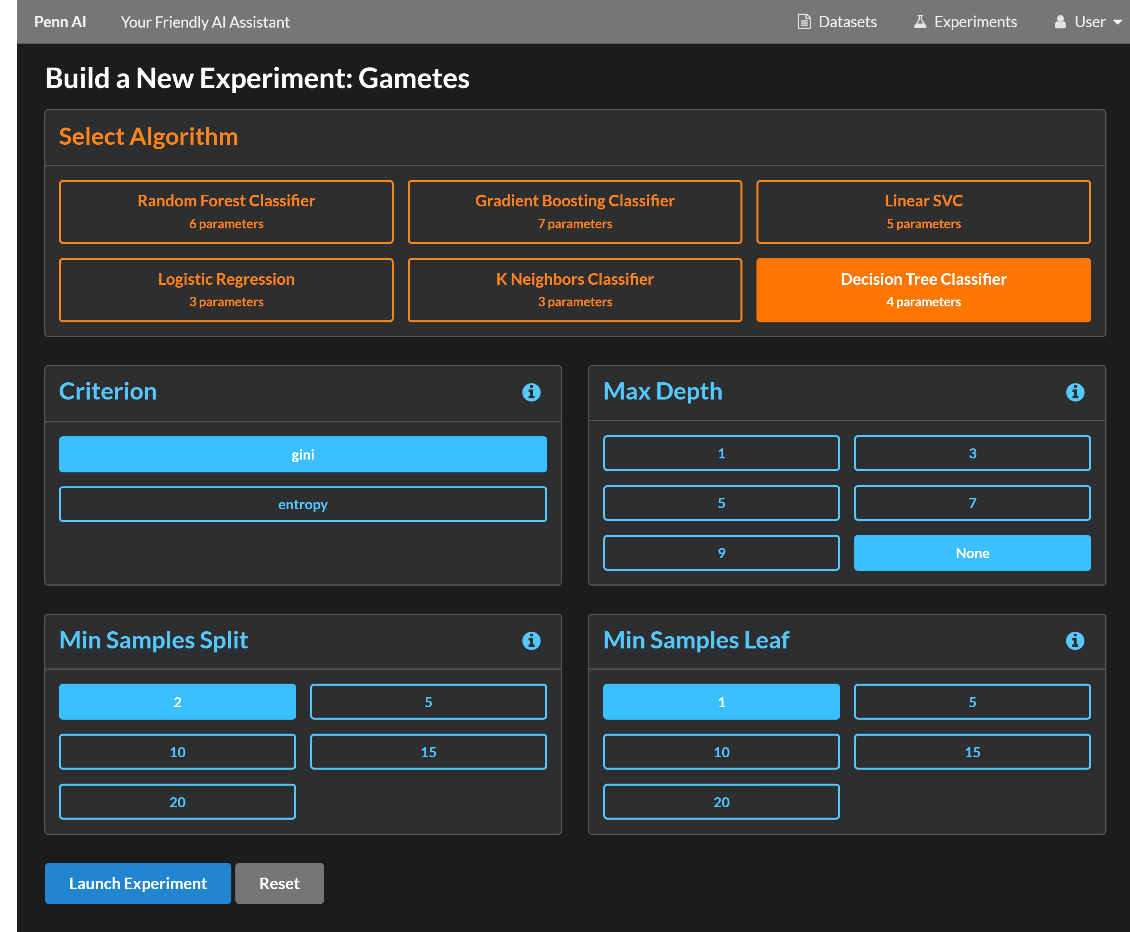}
\end{center}
\caption{Prototype of the Machine Learning Engine graphical user interface.}
\label{fig:prototype-ml-gui}
\end{figure}

Of course, there are dozens of machine learning algorithms, preprocessors, etc. to choose from in scikit-learn, which can be overwhelming to a novice user. To simplify the algorithm selection process for PennAI users, we currently limit PennAI to six machine learning algorithms that we believe will handle most supervised classification use cases, shown in Table~\ref{table:ml-algorithms}. We also limit the parameter choices for each algorithm to a handful of the most important parameters and parameter options, which makes it easier for users to choose a parameter configuration at the expense of algorithm customizability. An example of the interface to the Machine Learning Engine can be found in Figure~\ref{fig:prototype-ml-gui}, where only a handful of the most important parameters and parameter options are available for the k-Nearest Neighbors classification algorithm.

\begin{table}
    \centering
    \begin{tabular}{l l}
        \hline
        {\bf Classification} & {\bf Regression}\\ \hline
        Logistic Regression & ElasticNet\\
        Decision Tree & Decision Tree\\
        k-Nearest Neighbors & k-Nearest Neighbors\\
        Support Vector Machine & Support Vector Machine\\
        Random Forest & Random Forest\\
        Gradient Boosting & Gradient Boosting\\
        \hline
    \end{tabular}
    \caption{Machine learning algorithms available in PennAI.}
    \label{table:ml-algorithms}
\end{table}

In an upcoming PennAI implementation, we will provide simplified descriptions of the machine learning algorithms and parameters so users can make use of the algorithms without fully understanding their implementation. For example, when using a random forest it is not necessary for the user to understand what tuning the \texttt{n\_estimators} parameter does to the model. Instead, it is more important for the user to understand that adding more decision trees to the random forest (i.e., increasing \texttt{n\_estimators}) improves model performance but increases training time, whereas removing decision trees from the random forest decreases model performance but decreases training time~\cite{MachineLearningBook}.

Once the Machine Learning Engine finishes training and evaluating a machine learning model, it stores the machine learning model, the model predictions, and an analysis of the model in the Graph Database Engine, which are used in the Visualization Engine (both described below).

\section{The Controller Engine}

The Controller Engine acts as the interface between the high-performance computing system and the user or AI. This component is hidden from the user but facilitates the automatic launching of jobs on a multi-CPU machine, computing cluster, or cloud computing system. The controller must not only coordinate the launching of jobs but also keep track of when they finish and deposit the results in the Graph Database Engine (described below) that serves as the memory of the system.

For the Controller Engine, we selected an open source package called the Future Gadget Lab (FGLab), which is available on GitHub\footnote{FGLab: https://github.com/Kaixhin/FGLab}. FGLab functions as a server with individual runs launched as clients, called FGMachines. FGLab uses node.js to coordinate distributed jobs and uses MongoDB~\cite{Chodorow2010} as the backend database in the Graph Database Engine.

\section{The Graph Database Engine}

Another key component of PennAI is a memory system that keeps track of every analysis that is run on each data set. We keep track of the details of the machine learning method, the parameter settings, the data set analyzed, and results such as the model, model error, and area under the receiver operating characteristic curve (AUC). These are all stored in a JSON file that is deposited in a MongoDB NoSQL database. The advantage of using a NoSQL database is that new data elements can be added without creating tables and without strict format specifications. This flexibility is important for the rapidly changing landscape of machine learning. MongoDB can also function as a graph database that allows the documents to be linked in a network according to shared index terms related to the analysis and data. This feature facilitates more complex semantic queries of the database, such as ``Return the machine learning algorithm configurations that achieved the highest accuracy on any study involving prostate cancer.''

\subsection{Knowledge Base}
The Graph Database Engine serves as the memory of PennAI and provides the raw materials for the AI to learn which methods and parameter settings are working better than others for particular kinds of problems. The initial knowledge base consists of results from a previously published benchmark of scikit-learn algorithms~\cite{Olson2017PMLB}, in which 14 machine learning algorithms were run with full hyperparameter optimization on a suite of 165 supervised classification problems. The results are combined with meta-information about the datasets (e.g., number of features, number of instances, correlations between features, etc.) in order to allow the creation of a mapping from `problem instance space', i.e. dataset meta-features and model performance, to `learning space', i.e. machine learning algorithms and their parameters. This data can then be modelled to extract rules that represent the knowledge used by the Artificial Intelligence Engine to make informed analyses. The knowledge base will be updated with all future analyses.

\section{The Artificial Intelligence Engine}

Each component described above provides the raw materials for the Artificial Intelligence Engine which then 1) searches the graph database for results related to one or more data sets, 2) performs statistical analysis comparing algorithms and their parameters, 3) combines facts and rules in an expert system to make new analysis recommendations, 4) communicates findings to the user, and 5) automatically launches new analyses using suggestions from the expert system. The first function uses the search capabilities of the MongoDB graph database to identify relevant machine learning results in the form of JSON files. All returned JSON files can be parsed to extract the machine learning algorithm, parameters, and information about the model performance. These results are collated in a tab-delimited file and a statistical analysis performed to determine the best algorithm configurations for certain problem types, similar to meta-learning techniques~\cite{kalousis2002algorithm}.

New statistical results are used to populate the knowledge base of an expert system that has a set of decision rules provided by developers and advanced machine learning practitioners. This expert system is then used to make suggestions for additional analyses, for example by recommending better parameter settings or even entirely different machine learning algorithms that might be better-suited for the user's dataset. The user can access these suggestions manually or PennAI can use the suggestions to automatically launch new jobs, thus continually growing the PennAI knowledge base. Essentially, the Artificial Intelligence Engine becomes a research assistant who tinkers with new ways of modeling the dataset and reports back to the user with their best findings.

\section{The Visualization Engine}

\begin{figure}
\begin{center}
\includegraphics[width=\textwidth]{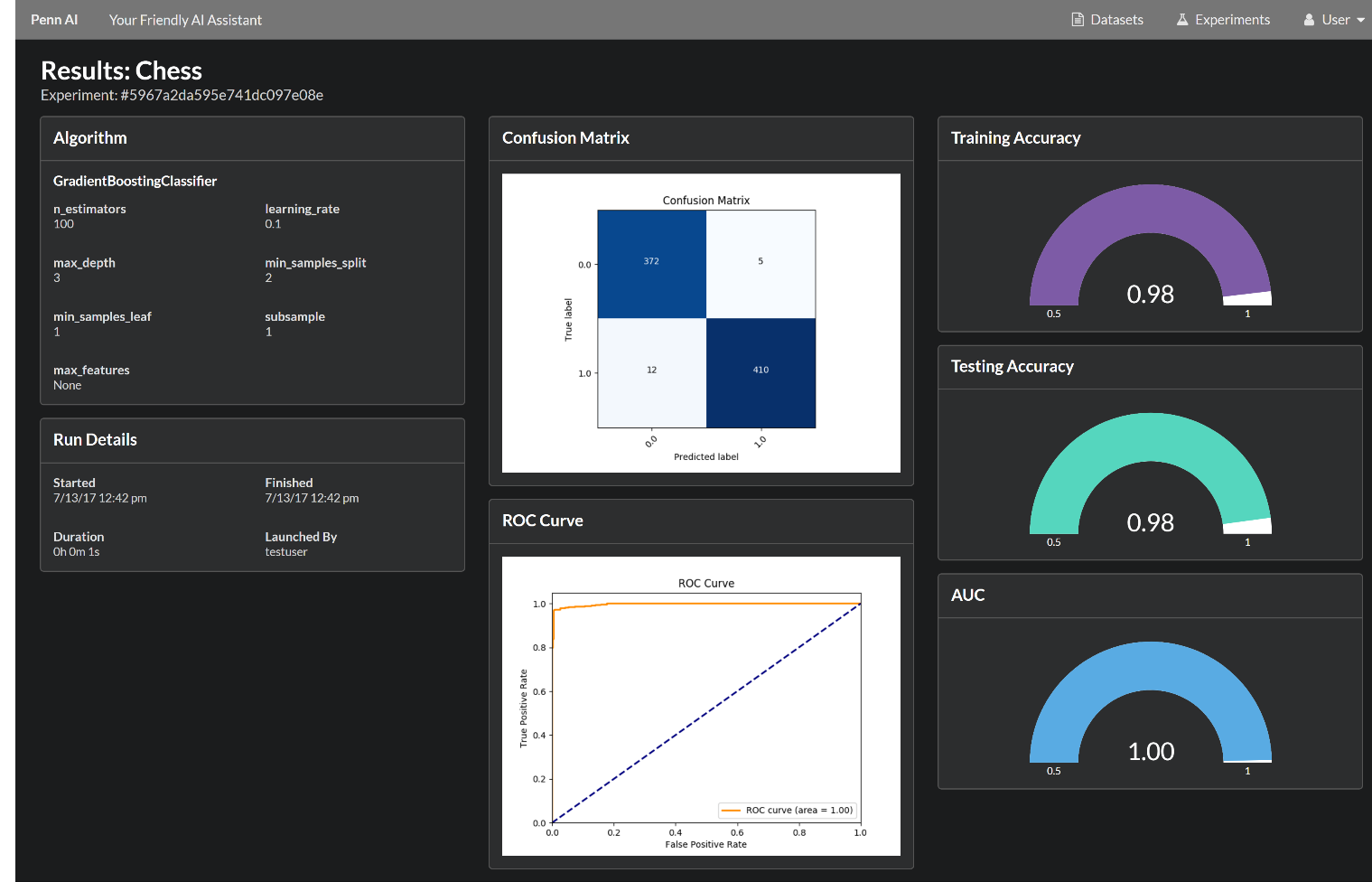}
\end{center}
\caption{Prototype of the Visualization Engine graphical user interface.}
\label{fig:prototype-visualization-engine}
\end{figure}

Visualization will be critical for fostering the human-AI collaboration described above. The user will need to be able to see individual machine learning models and results as well as higher-level results from statistical analyses across machine learning runs. We extract visual results such as the receiver operating characteristic (ROC) curves and models to store in the graph database, as shown in Figure~\ref{fig:prototype-visualization-engine}. PennAI will also generate heatmaps and other visualizations that summarize results across different machine learning methods and datasets. These higher-level visualizations will aid the user with making decisions about new manual analyses to launch and will help them assess how well the PennAI assistant is doing. These images will be linked to the datasets and results in the Graph Database Engine, and will thus be easily searchable.

\section{Discussion and Future Work}

Thus far, we have described PennAI as a system that provides a simple interface for users to upload their datasets, launch machine learning analyses, view the results of the analyses in an intuitive manner, and use those results to refine their machine learning analyses. We also described how PennAI will use a combination of expert knowledge from advanced machine learning practitioners and prior statistical knowledge of machine learning algorithm performance on datasets to recommend new analyses to the user, as well as launch its own analyses to later report to the user. In essence, the primary goal of PennAI is to provide an AI research assistant for its users. However, considering the name of this workshop and book---Genetic Programming Theory and Practice---one may be left wondering how GP can be incorporated into PennAI. In the following paragraphs, we will describe our plans for integrating GP into PennAI.

The first point of entry is to include GP as a machine learning option since a number of successful biomedical applications have been reported (e.g., ~\cite{vanneschi2009classification, moore2015identification,moore2008development,Moore2011,Moore2015,Moore2014}). A GP system for classification based on multidimensional clustering~\cite{silva2016multiclass} was recently demonstrated on biomedical classification problems~\cite{la2017genetic} as a competitive alternative to traditional machine learning approaches. Recently GP has been proposed as a general feature engineering wrapper (FEW)\footnote{http://lacava.github.io/few} in order to harness its feature learning capability to improve scikit-learn estimators, both for regression~\cite{la2017general} and classification~\cite{la2017ensemble}. FEW allows GP to provide readable feature transformations to users while still utilizing existing modeling techniques for making predictions. As mentioned in Section 2, interpretation and explanatory power are important aspects of using AI for data mining, and therefore GP methods that produce concise models, e.g. by local search~\cite{la2016inference} or Pareto optimization~\cite{la2016automatic}, are important options to include. Further down the road, it could be possible for PennAI to allow advanced users to incorporate custom machine learning algorithms into PennAI by providing a scikit-learn formatted interface to their project (e.g. ellyn\footnote{http://epistasislab.github.io/ellyn}). PennAI could then provide a ``bring your own learner" type of service~\cite{arnaldo2015bring} to allow researchers to tackle complex data mining tasks with customized learning approaches, and incorporate the results into its knowledge base for improving future data science projects.   

Beyond using GP to perform the machine learning itself, recent work has shown that GP can also be harnessed to optimize a sequence of existing data analysis and machine learning operations on a dataset to maximize the predictive performance of the final machine learning model~\cite{zutty2015multiple,de2017recipe}. For example, TPOT\footnote{https://github.com/rhiever/tpot} is an early prototype that uses GP to optimize a sequence of scikit-learn operations for both classification and regression problems~\cite{Olson2016EvoBio,Olson2016JMLR,Olson2016GPTP}, and has been shown to work quite well across a broad range of application domains ranging from epidemiological studies to image classification to time series prediction~\cite{OlsonGECCO2016}. Given the general design of TPOT, the operations it optimizes over can be specialized for particular problem domains. As another example, the TPOT-MDR project~\cite{Sohn2017GECCO} showed that TPOT can be specialized for genome-wide association studies (GWAS), and it outperforms several state-of-the-art modeling methods on both simulated and real-world GWAS problems because it considers a broad range of operations in with one another. As such, we view GP as a strong candidate for a future version of the PennAI Artificial Intelligence Engine, where the GP is seeded with the best known algorithm configurations and uses the core principles of GP (inheritance, mutation, and crossover)---distributed over a high-performance computing cluster---to improve the algorithm configurations from there. This brand of GP-based AI system would be useful for automatically launching new analyses, but less useful for recommending particular algorithm configurations to the user because GP does not provide a notion of the ``next best'' solution to attempt.

Another offshoot of PennAI we are currently investigating is the use of a meta genetic algorithm to find parameters (population size, generation count, etc.) for a GP instance that work well, i.e., solve a given problem~\cite{sipper2017metaga}. This meshes well with the idea that the AI of PennAI will aid non-machine-learning experts run complex algorithms, such as GP, without having to find or even understand every single parameter. 

Ultimately, PennAI will likely be comprised of several disparate AI algorithms that use meta-data and meta-learning to improve the user experience and user productivity by suggesting machine learning algorithms and parameters, as well as providing other insights. As a result, we will be able to harness ensemble techniques to collate the advice given by the numerous AI algorithms.

The time is now to bring AI technology to anyone that wants to use it for big data analytics. The software and hardware technology exists and data has never been bigger, more complex, and more plentiful. PennAI will provide both machine learning and AI capability to both naive and expert users alike with a user-friendly web and smartphone-enabled interface. We see AI technology such as PennAI not as a replacement for the data scientist but rather as a data science assistant that can suggest analyses to the user or provide automatically generated results that are informed by previous analyses across different data sets. The user can take these results as-is or use them as inspiration in manual analyses. The democratization of AI is here.

\section{Acknowledgements}

This work was generously funded provided by the Perelman School of Medicine and the University of Pennsylvania Health System of the University of Pennsylvania. Additional funding was provided by National Institutes of Health grants AI116794, DK112217, ES013508, and TR001878.

\bibliography{bib}{}
\bibliographystyle{spmpsci}

\end{document}